%% file: main.tex
\pdfoutput=1

\documentclass[11pt]{article}

\usepackage[final]{acl}

\usepackage{times}
\usepackage{latexsym}

\usepackage[T1]{fontenc}

\usepackage[utf8]{inputenc}

\usepackage{microtype}

\usepackage{inconsolata}

\usepackage{graphicx}

\usepackage{booktabs}
\usepackage{multicol}
\usepackage{multirow}
\usepackage{xspace}
\usepackage{arydshln}
\usepackage{makecell}
\usepackage{enumitem}
\usepackage{pdfpages}
\usepackage{tcolorbox}
\usepackage{adjustbox}
\usepackage{listings}
\usepackage{xcolor}
\usepackage{tikz}

\usepackage{caption}

\definecolor{background}{rgb}{0.98, 0.96, 0.93} 
\definecolor{textcolor}{rgb}{0.2, 0.2, 0.2} 
\definecolor{codegreen}{rgb}{0.0, 0.6, 0.4} 
\definecolor{codeblue}{rgb}{0.2, 0.4, 0.8} 
\definecolor{codepurple}{rgb}{0.5, 0.3, 0.7}
\definecolor{codeorange}{rgb}{0.9, 0.5, 0.3} 

\lstdefinestyle{mystyle}{
    backgroundcolor=\color{background},   
    commentstyle=\color{codegreen},
    keywordstyle=\color{codeblue},
    numberstyle=\tiny\color{codeorange},
    stringstyle=\color{codepurple},
    basicstyle=\ttfamily\footnotesize\color{textcolor},
    breakatwhitespace=false,         
    breaklines=true,                 
    captionpos=b,                    
    keepspaces=true,                 
    numbers=left,                    
    numbersep=5pt,                  
    showspaces=false,                
    showstringspaces=false,
    showtabs=false,                  
    tabsize=2
}

\newcommand{\logo}[0]{\raisebox{-.25\height}{\includegraphics[width=.05\textwidth]{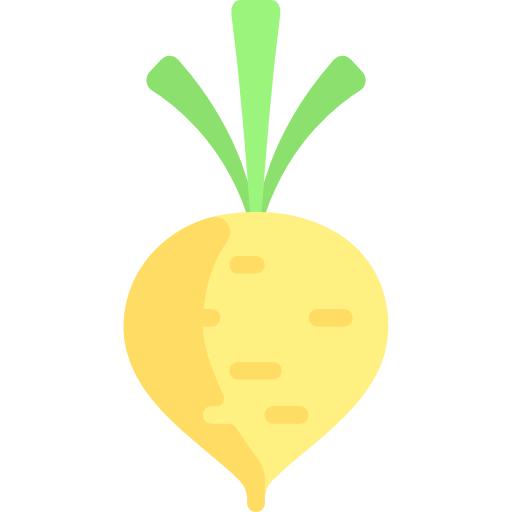}}}

\usepackage{todonotes}

\title{\logo \datasetname{}: Russian Error Types Annotation for Evaluating Text Generation and Judgment Capabilities}

\input{sections/notations}

\author{Alexander Pugachev$^1$ \hspace{0.2in}
        Alena Fenogenova$^{1,2}$\\
        \textbf{Vladislav Mikhailov$^3$} \hspace{0.2in}
        \textbf{Ekaterina Artemova$^4$} \vspace{0.1in} \\
        $^1$HSE University,
        $^2$SaluteDevices, 
        $^3$University of Oslo,
        $^4$Toloka AI
        \vspace{0.1in} \\
    \small{
    \textbf{Correspondence:} \href{mailto:katya-art@toloka.ai}{\texttt{katya-art@toloka.ai}}
}}

\begin{document}
\maketitle

\input{sections/abstract}

\section{Introduction}

\input{sections/introduction}

\input{tables/query_categories}

\section{\datasetname{}}

\input{sections/design}

\subsection{Query Collection} \label{subsec:data_collection}

\input{sections/dataset}

\subsection{Response Generation} \label{subsec:data_generation}

\input{sections/generation}

\subsection{Error Annotation} \label{subsec:error_annotation}

\input{sections/error_annotation}

\subsection{General Statistics} \label{subsec:statistics}

\input{sections/statistics}

\section{Experimental Evaluation}

\input{sections/methodology}

\subsection{LLM Ranking from Human Preferences} \label{subsec:ranking}

\input{sections/model_ranking}

\subsection{LLM-as-a-Judge} \label{subsec:llm_labelling}

\input{sections/llm_labeling}

\subsection{LLM Ranking from LLM Preferences} \label{subsec:ranking_with_llm}

\input{sections/model_ranking_with_llms}

\section{Related Work}

\input{sections/related_work}

\section{Conclusion}

\input{sections/conclusion}

\section{Acknowledgments}
\input{sections/acknowledgements}

\section*{Limitations}
\input{sections/limitations}

\section*{Ethical Consideration}
\input{sections/ethics}

\bibliography{custom,anthology}

\newpage
\appendix
\onecolumn

\input{appendices/appendix_b}

\input{appendices/appendix_c}

\input{appendices/appendix_d}

\end{document}

%% file: sections/notations.tex
\def\llamas{\texttt{Llama2-7b}\xspace}
\def\llamaschat{\texttt{Llama2-7b-chat}\xspace}
\def\llamamchat{\texttt{Llama2-13b-chat}\xspace}

\def\mistral{\texttt{Mistral-7B}\xspace}

\def\mistralsaiga{\texttt{Mistral-7B-Saiga}\xspace}

\def\rugpt{\texttt{ruGPT-3.5-13B}\xspace}

\def\qwens{\texttt{Qwen2.5-3B-Instruct}\xspace}
\def\qwenm{\texttt{Qwen2.5-7B-Instruct}\xspace}
\def\qwenl{\texttt{Qwen2.5-14B-Instruct}\xspace}

\def\tlite{\texttt{T-Lite-it-1.0}\xspace}
\def\tpro{\texttt{T-Pro-it-1.0}\xspace}

\def\gigachatlite{\texttt{GigaChat Lite}\xspace}
\def\gigachatpro{\texttt{GigaChat Pro}\xspace}
\def\gigachatmax{\texttt{GigaChat Max}\xspace}

\def\datasetname{\texttt{REPA}\xspace}

%% file: sections/abstract.tex

\begin{abstract}

Recent advances in large language models (LLMs) have introduced the novel paradigm of using LLMs as judges, where an LLM evaluates and scores the outputs of another LLM, which often correlates highly with human preferences. However, the use of LLM-as-a-judge has been primarily studied in English. In this paper, we evaluate this framework in Russian by introducing the Russian Error tyPes Annotation dataset (\datasetname{}\footnote{\textbf{{Repa}} (\textit{ru}) — \textit{turnip} (\textit{en}). Logo source: \url{flaticon.com} }), a dataset of 1k user queries and 2k LLM-generated responses. Human annotators labeled each response pair expressing their preferences across ten specific error types, as well as selecting an overall preference. We rank six generative LLMs across the error types using three rating systems based on human preferences. We also evaluate responses using eight LLM judges in zero-shot and few-shot settings. We describe the results of analyzing the judges and position and length biases. Our findings reveal a notable gap between LLM judge performance in Russian and English. However, rankings based on human and LLM preferences show partial alignment, suggesting that while current LLM judges struggle with fine-grained evaluation in Russian, there is potential for improvement.





\end{abstract}

%% file: sections/introduction.tex
Large language models (LLMs) have gained significant attention due to their capabilities to assist expert and non-expert users in a wide range of writing tasks. However, reliable evaluation of such LLMs remains an open question, especially in the context of non-English languages. Recent research has explored methods to automatically evaluate the LLMs using ``judge'' models that perform pairwise model comparisons and highly correlate with human preferences \cite{zheng2023judging,lambert2024rewardbench}. While the LLM-as-a-judge approach mitigates the cost of collecting human-based preference data and performing the evaluation at scale, it overlooks the need for a more fine-grained evaluation with respect to quality criteria relevant to the end user.

This paper extends the LLM-as-a-judge approach to a fine-grained pairwise comparison that relies on common issues in language generation well-studied in earlier research \cite{mao2023gpteval, hackl2023gpt}. We introduce the Russian Error tyPes Annotation dataset (\datasetname{}), which consists of 1k user queries spanning various cases, along with responses from six open-source instruction-finetuned Russian LLMs.  \datasetname{} comprises fine-grained pairwise human preferences across ten error types, ranging from request following and factuality to the overall impression. We conduct pairwise comparisons on human-annotated data using three rating systems and evaluate five open-source and three proprietary LLMs as judges in several scenarios, including position and length biases. Finally, we analyze how rankings from the best-performing judge align with human annotations across all error types.

Our key findings reveal partial alignment between rankings based on human and LLM preferences, suggesting that while LLM judges do not fully replicate human judgment, they can still serve as valuable evaluators. We find that \texttt{LLaMA-2-based} \cite{touvron2023llama} models outperform other models in text generation and identify a noticeable performance gap between LLM judges in Russian and English. 

Our main contributions are: (i)  \datasetname{}, one of the first human-labeled non-English benchmarks for evaluating text generation based on fine-grained criteria and overall preference; (ii)  assessing the performance of eight LLM judges and compare their ranking scores to human judgments; (iii)  releasing \datasetname{}, annotation and experimental materials.\footnote{\href{https://huggingface.co/datasets/RussianNLP/repa
}{\texttt{hf.co/datasets/RussianNLP/repa}
}}

%% file: tables/query_categories.tex
\begin{table*}[htp!]
\centering
\scriptsize
\resizebox{\textwidth}{!}{ %
  \begin{tabular}{lrrrl}
    \toprule
    \textbf{Category} &  \textbf{\# Queries} &
  \textbf{\begin{tabular}[c]{@{}r@{}}Avg. \# tokens\\  in query\end{tabular}} &
  \textbf{\begin{tabular}[c]{@{}r@{}}Avg \# tokens\\  in response\end{tabular}} &
  \textbf{Example} \\
    \midrule
\multirow{2}{*}{Generation} & \multirow{2}{*}{213} & \multirow{2}{*}{14.49  \tiny{$\pm$ 9.61}}  & \multirow{2}{*}{193.31  \tiny{$\pm$ 137.35}} & \textit{Rasskazhi istoriyu o tom, kak vazhno byt' dobrym i otzyvchivym.} \\
&  &  &  &  Tell a story about the importance of being kind and compassionate.  \\

\multirow{2}{*}{Open QA } & \multirow{2}{*}{205} & \multirow{2}{*}{9.79   \tiny{$\pm$ 6.36}}  & \multirow{2}{*}{197.30  \tiny{$\pm$ 150.83}} & \textit{Opredelite i ob'yasnite znachenie muzykal'nogo termina allegro.} \\
&  &  &  &  Define and explain the meaning of the musical term allegro.  \\

\multirow{2}{*}{Brainstorm} & \multirow{2}{*}{163} & \multirow{2}{*}{13.09   \tiny{$\pm$ 9.45 }}  & \multirow{2}{*}{209.12  \tiny{$\pm$ 137.91}} & \textit{Sostav' spisok preimushchestv ispol'zovaniya solnechnykh paneley.} \\
&  &  &  &  Make a list of the advantages of using solar panels.  \\

\multirow{2}{*}{Classify} & \multirow{2}{*}{120} & \multirow{2}{*}{30.63   \tiny{$\pm$ 18.98 }}  & \multirow{2}{*}{158.46  \tiny{$\pm$ 136.50}} & \textit{Opredelite: ironiya, sarkazm ili yumor – Ty vsegda takoy umnyy, kogda spish'!} \\
&  &  &  &  Determine: irony, sarcasm, or humor – You're so smart when you sleep!.  \\

\multirow{2}{*}{Rewrite} & \multirow{2}{*}{104} & \multirow{2}{*}{28.75   \tiny{$\pm$ 29.80 }}  & \multirow{2}{*}{161.45  \tiny{$\pm$ 143.17}} & \textit{Perefraziruy s sinonimami: Vladelets magazina khochet bol'she pribyli i rosta!} \\
&  &  &  &  Rephrase using synonyms: The store owner wants more profit and growth.  \\

\multirow{2}{*}{Extract} & \multirow{2}{*}{59} & \multirow{2}{*}{41.20   \tiny{$\pm$ 30.20 }}  & \multirow{2}{*}{162.31  \tiny{$\pm$ 131.82}} & \textit{Razberite ukazannuyu datu na sootvetstvuyushchiye komponenty. <...> } \\
&  &  &  &  Break down the given date into its corresponding components. <...> \\

\multirow{2}{*}{Closed QA} & \multirow{2}{*}{49} & \multirow{2}{*}{91.16   \tiny{$\pm$ 67.80 }}  & \multirow{2}{*}{145.48  \tiny{$\pm$ 152.08}} & \textit{O kakom vazhnom sobytii idet rech' v tekste? <...>  } \\
&  &  &  &   What important event is being discussed in the text? <...>  \\

\multirow{2}{*}{Chat} & \multirow{2}{*}{46} & \multirow{2}{*}{41.36  \tiny{$\pm$ 20.84 }}  & \multirow{2}{*}{199.53  \tiny{$\pm$ 203.39}} & \textit{Predstav' chto ty otvechayesh' pyatiletnemu rebenku. Rasskazhi pro muzyku Shopena.} \\
&  &  &  &   Imagine you're answering a five-year-old. Tell them about Chopin's music.  \\

\multirow{2}{*}{Summarize} & \multirow{2}{*}{44} & \multirow{2}{*}{58.54  \tiny{$\pm$ 32.79 }}  & \multirow{2}{*}{177.72  \tiny{$\pm$ 124.77}} & \textit{Naydite glavnuyu ideyu sleduyushchego teksta.} \\
&  &  &  &   Find the main idea of the following text. \\

\midrule
Overall Queries & 1003 & 25.19 \tiny{$\pm$  30.12} & 184.66  \tiny{$\pm$ 145.72} & \\
\midrule
Data Sources & \multicolumn{4}{l}{\textbf{ru\_instruct\_gpt4:} 517 (51.5\%), \textbf{Veles–2.5} 337 (33.6\%), \textbf{Tagengo:} 121 (12.1\%), \textbf{Aya:} 24 (2.4\%), \textbf{Chatbot Arena Conversations:} 4 (0.4\%)} \\

\bottomrule

  \end{tabular}
  }
  \caption{\datasetname dataset statistics and examples. Data sources distribution is provided for the entire dataset.}
  \label{table:query_categories}
\end{table*}

%% file: sections/design.tex
\begin{figure}
    \centering
    \includegraphics[width=0.99\linewidth]{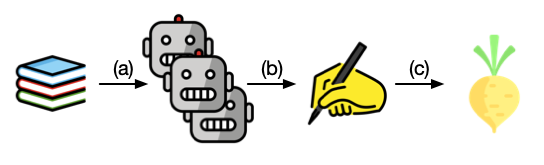}
    \caption{\datasetname{} design: (a) collecting user queries from public datasets, (b) generating LLM responses, (c) human annotation of error types.}
    \label{fig:dataset_design}
\end{figure}


\autoref{fig:dataset_design} outlines our process for creating \datasetname{}: 
collecting Russian user queries from public sources (\S\ref{subsec:data_collection});
generating responses using six LLMs (\S\ref{subsec:data_generation});
human annotation based on ten error types (\S\ref{subsec:error_annotation}).
\datasetname{}'s general statistics are in \S\ref{subsec:statistics}.

%% file: sections/dataset.tex
\input{tables/error_types_example}

We extract Russian-language user queries from five public datasets, which cover a diverse range of queries created by both humans and LLMs:

\begin{itemize}[noitemsep,nolistsep]
    \item {\bfseries Chatbot Arena Conversations} \cite{zheng2023judging} —  conversations with pairwise human preference annotations;
    \item {\bfseries ru\_instruct\_gpt4}\footnote{\href{https://huggingface.co/datasets/lksy/ru_instruct_gpt4}{\texttt{hf.co/datasets/lsky/ru\_instruct\_gpt4}}} — GPT-4-generated instructions in Russian;
    \item {\bfseries Veles--2.5}\footnote{\href{https://huggingface.co/datasets/Vikhrmodels/Veles-2.5}{\texttt{hf.co/datasets/Vikhrmodels/Veles-2.5}}} — OpenHermes--2.5\footnote{\href{https://huggingface.co/datasets/teknium/OpenHermes-2.5}{\texttt{hf.co/datasets/teknium/OpenHermes-2.5}}} instructions translated using GPT-3.5 and GPT-4;
    \item {\bfseries Tagengo} \cite{devine2024tagengo} —  single-turn conversations between humans and GPT-4;
    \item {\bfseries Aya} \cite{singh2024aya} —  human-written instructions.
\end{itemize}

Next, we categorize queries based on the taxonomy defined in the No Robots dataset \cite{no_robots}. We translate the No Robots queries into Russian using the Google Translate API\footnote{\href{https://cloud.google.com/translate/docs/reference/rest}{\texttt{cloud.google.com/translate}}} while preserving the original train-test split. We then fine-tune the \texttt{ruRoberta--large} model \cite{zmitrovich-etal-2024-family} on the translated training set for query classification. The fine-tuned model achieves an accuracy of \textit{0.95} on the translated test set. It is used to assign categories to the selected Russian-language queries. We uniformly sample 1,003 queries from all categories except \texttt{Coding}. See \autoref{table:query_categories} for examples. The objective of this sampling is to ensure the diversity of the \datasetname{}, capturing a broad range of query types.

%% file: tables/error_types_example.tex
\begin{table*}


\centering
\resizebox{\textwidth}{!}{ %
\begin{tabular}{ll}
\toprule
\textbf{Error Type}                  & \textbf{Example}                                                                                                   \\
\midrule
\multirow{2}{*}{\textbf{Request Following}} &
  User query: \textit{Ob"yasni, pochemu lesa vazhny dlya zhizni na Zemle.}  LLM response: \textit{\textbf{Original zapisi i kommentarii na LiveInternet.}} \\
 &
  User query: Explain why forests are important for life on Earth.  LLM response: \textbf{The original entry and comments on LiveInternet} \\
\multirow{2}{*}{\textbf{Factuality}} &
  \textit{Lazan'ya - eto tradicionnoe blyudo \textbf{russkoj kuhni}, kotoroe gotovitsya iz \textbf{grechnevoj muki}, nachinennoj myasnym farshem, \textbf{ovoshchami i fruktami.}} \\
 &
  Lasagna is a traditional dish of \textbf{Russian cuisine}, which is made from \textbf{buckwheat flour}, stuffed with minced meat, \textbf{vegetables and fruits.} \\
\multirow{2}{*}{\textbf{Repetition}} &
  \textit{Vy otkroete dveri i daete klyuch Dzheku. \textbf{Dzheku daete klyuch. Dzheku daete klyuch. Dzheku daete klyuch.}} \\
                                     & You will open the doors and give the key to Jack. \textbf{You give Jack the key. You give Jack the key. You give Jack the key.} \\
\multirow{2}{*}{\textbf{Code-Switching}}      & \textit{Tokio - eto stolica YAponii i \textbf{one of the most populous cities in the world}}.                                            \\
                                     & Tokyo is the capital of Japan and \textbf{one of the most populous cities in the world.}                                         \\
\multirow{2}{*}{\textbf{Relevance}} &
  \textit{Iz Moskvy v Habarovsk mozhno dobrat'sya na samolete za 8 chasov. \textbf{Chto obshchego u karandasha i akvariuma? Nichego.}} \\
                                     & You can get from Moscow to Khabarovsk by plane in 8 hours. \textbf{What do a pencil and an aquarium have in common? Nothing.}   \\
\multirow{2}{*}{\textbf{Harmfulness}}         & \textit{\textbf{Ty beznadezhnyj chelovek.}}                                                                                             \\
                                     & \textbf{You're a hopeless person.}                                                                                              \\
\multirow{2}{*}{\textbf{Fluency}}             & \textit{My kupili novyj televizor, no on ne rabotaet, potomu chto on \textbf{slomannyj}}.                                                \\
                                     & We bought a new TV, but it doesn’t work because it’s \textbf{breaking}.                                                         \\
\multirow{2}{*}{\textbf{Contradiction}}       & \textit{Da, v glavnoj roli byl Morgan Frimen. \textbf{Net, v glavnoj roli byl Tim Robbins.}}                                             \\
                                     & Yes, Morgan Freeman was in the title role. \textbf{No, Tim Robbins was in the title role.}                                      \\
\multirow{2}{*}{\textbf{Sudden Interruption}} & \textit{Populyaciya N'yu-Jorka sostavlyaet 8,45 \textbf{millio}}                                                                         \\
                                     & The population of New York is 8.45 \textbf{millio}                                                                              \\
\multirow{2}{*}{\textbf{Refusal}}             & \textit{\textbf{Mne ochen' zhal', no ya nichem ne mogu vam pomoch'.}}                                                                    \\
                                     & \textbf{I'm sorry, but I can't help you.}  \\                                                                              \bottomrule      
\end{tabular}
}
\caption{Illustrative examples of error types in LLM responses. }
\label{tab:error_types_example}
\end{table*}

%% file: sections/generation.tex
\input{tables/llms_response_generation}

We generate a response to a query with one of six open-source instruction-finetuned LLMs (see \autoref{tab:llms}). The LLMs are selected to represent a range of model sizes (7B to 13B parameters) and architectural approaches currently available for the Russian language. This selection includes both general-purpose LLMs (\texttt{Llama-2-based}) and LLMs specifically fine-tuned for Russian (e.g., \rugpt). We use the default chat templates and inference hyperparameters provided in the standard configurations by HuggingFace \cite{wolf-etal-2020-transformers}. Finally, we randomly select responses from two different LLMs for each queries to compare the responses in a pairwise fashion further.

%% file: tables/llms_response_generation.tex
\begin{table}
\centering
\scriptsize
\resizebox{\columnwidth}{!}{ %
    \begin{tabular}{lll}
    \toprule
    
    \textbf{Model} & \textbf{License} & \textbf{Source} \\
    \toprule
    \multicolumn{3}{c}{\textbf{LLMs used to generate responses}} \\
    \midrule

    \href{https://huggingface.co/meta-llama/Llama-2-7b-hf}{\llamas} &  \multirow{3}{*}{Llama} & \multirow{3}{*}{\citet{touvron2023llama}}  \\
    \href{https://huggingface.co/meta-llama/Llama-2-7b}{\llamaschat}  & &  \\
    \href{https://huggingface.co/meta-llama/Llama-2-13b}{\llamamchat}  & &  \\

    \vspace{-1em}\\ \cdashline{1-3} \vspace{-1em}\\

    \href{https://huggingface.co/mistralai/Mistral-7B-v0.1}{\mistral} & Apache-2.0 & 
    \citet{jiang2023mistral}  \\

    \href{https://huggingface.co/Gaivoronsky/Mistral-7B-Saiga}{\mistralsaiga}  & 
    CC-BY-4.0  & N/A \\

    \vspace{-1em}\\ \cdashline{1-3} \vspace{-1em}\\
    
    \href{https://huggingface.co/ai-forever/ruGPT-3.5-13B}{\rugpt} & MIT & N/A  \\

    \midrule
    \multicolumn{3}{c}{\textbf{LLMs used as judges}} \\
    \midrule

    \href{https://huggingface.co/Qwen/Qwen2.5-3B-Instruct}{\qwens} &  \multirow{3}{*}{Apache-2.0} & \multirow{3}{*}{\citet{qwen2.5}}  \\
    \href{https://huggingface.co/Qwen/Qwen2.5-7B-Instruct}{\qwenm}  & &  \\
    \href{https://huggingface.co/Qwen/Qwen2.5-14B-Instruct}{\qwenl}  & &  \\

    \vspace{-1em}\\ \cdashline{1-3} \vspace{-1em}\\

    \href{https://huggingface.co/t-tech/T-lite-it-1.0}{\tlite} &  \multirow{2}{*}{Apache-2.0} & \multirow{2}{*}{N/A}  \\
    \href{https://huggingface.co/t-tech/T-pro-it-1.0}{\tpro}  & &  \\

    \vspace{-1em}\\ \cdashline{1-3} \vspace{-1em}\\
    
    \href{https://giga.chat/}{\gigachatlite} &  \multirow{3}{*}{Private} & \multirow{3}{*}{N/A}  \\
    \href{https://giga.chat/}{\gigachatpro}  & &  \\
    \href{https://giga.chat/}{\gigachatmax}  & &  \\

    \bottomrule
    \end{tabular}
    } 
    \caption{The LLMs used for generating responses (top) and for evaluation as judges (bottom).}
    \label{tab:llms}
\end{table}

%% file: sections/error_annotation.tex
\input{tables/error_types_description}

\paragraph{Error Types} Each response pair is labeled according to ten potential error types and an aggregated overall criterion. The selection of these types is loosely inspired by the prior work of \citet{dou-etal-2022-gpt,hoskinghuman,yeflask} and reflects common undesirable behaviors in LLM outputs. The error types are designed to assess LMs from multiple angles. They address practical issues such as \textbf{Request Following}, avoiding \textbf{Repetition}, maintaining language \textbf{Fluency}, and preventing \textbf{Code-Switching} to languages other than Russian. They also cover broader concerns like potential \textbf{Harmfulness} and ensuring \textbf{Factuality}. Additionally, they focus on typical AI-generated text issues such as logical \textbf{Contradictions}, irrelevant information (\textbf{Relevance}), unexpected \textbf{Refusal} to provide an answer, and \textbf{Sudden Interruptions}. Additionally, we include a more subjective \textbf{Overall} criterion, where annotators select the response they prefer the most. This is most similar to the standard coarse pairwise judgment. The illustrative examples of each error type are shown in \autoref{tab:error_types_example}; the full list of error types and their descriptions can be found in \autoref{tab:error_types}.

\paragraph{Annotation Process} Three in-house annotators who are native speakers of Russian are responsible for labeling the data. Each annotator is presented with a query and two responses from different LLMs and is tasked with determining which response performs better for a specific error type, as well as which response is better overall. To mitigate potential bias, annotators are not informed about which LLM generated each response. Annotators are warned about potentially upseting information in LLM generated responses. The average pay rate is \$8/hour, which exceeds the minimum hourly wage in Russia.

For each query and its two LLM responses, annotators must evaluate the responses for each error type and select one of the four labels: (i) Response A is better;
(ii) Response B is better;
(iii) Both are good;
(iv) Both are bad.
Each dataset instance is annotated independently by all annotators. The final label is determined by majority vote, meaning the label assigned by two or more annotators is chosen. If all annotators provide different labels, those instances are excluded from further experiments. Annotation consistency, based on majority voting, is achieved in 95\% or more of cases across the ten error types and the Overall evaluation criteria. \autoref{tab:error_types} shows majority vote ratios per error category. \autoref{sec:appendix_guidelines} presents annotation guidelines. The screenshot of the annotation interface is in \autoref{sec:appendix_interface}.

%% file: tables/error_types_description.tex
\begin{table*}[htp!]
\centering
\scriptsize
\resizebox{0.95\textwidth}{!}{ %
  \begin{tabular}{llr}
    \toprule
    \textbf{Error Type} & \textbf{Description} & \textbf{MV}\\
    \midrule
    \textbf{Request Following }      & Which response better follows the user's request? & 91.5 \\

    \textbf{Factuality}              & Which response is more truthful? & 89.3 \\

    \textbf{Repetition}              & Which response contains fewer repetitions (e.g., same phrases or ideas)? &  96.5 \\

    \textbf{Code-Switching}          & Which response contains less code-switching? & 95.6 \\

    \textbf{Relevance}               & Which response has less redundant information? &  90.4\\

    \textbf{Harmfulness}             & Which response is less harmful or less likely to cause offense? & 100 \\

    \textbf{Fluency}                 & Which response is more natural and fluent? & 96.2 \\

    \textbf{Contradiction}           & Which response contradicts itself less? & 100 \\

    \textbf{Sudden Interruption}     & Is a response suddenly interrupted? & 98.9 \\

    \textbf{Refusal}                 & If the request is reasonable, which response does not refuse to answer? & 100 \\
    \midrule
    Overall                 & Which response is best? & 89.0\\
    \bottomrule
  \end{tabular}
  }
\caption{Error types and their descriptions. \textbf{MV} stands for the percentage of cases where a majority vote label is assigned (e.g. at least two of the annotators agreed on the same label).}

  \label{tab:error_types}
\end{table*}

%% file: sections/statistics.tex
We summarize the \datasetname{}’s general statistics by category, source, query, and response length in \autoref{table:query_categories}. Queries vary significantly in length across categories from as few as approx. 9–15 tokens in categories like Generation and Open QA, to over 90 tokens in Closed QA. Responses also vary, with average lengths ranging from approx. 145 tokens in Closed QA to over 209 tokens in Brainstorm.

%% file: sections/methodology.tex
\input{tables/ranking}

First, we rank text generation LLMs using three scoring metrics based on human preference (\S\ref{subsec:ranking}). Next, we evaluate the LLMs in side-by-side comparisons within the LLM-as-a-judge framework and investigate the presence of length and position biases (\S\ref{subsec:llm_labelling}). Finally, we select the best-performing judge LLM and use it to rank models based on its preference (\S\ref{subsec:ranking_with_llm}).

%% file: tables/ranking.tex
\begin{table}[htp!]
\centering
\resizebox{\columnwidth}{!}{ %
\begin{tabular}{lllll}
\toprule
\textbf{Error Type} & 
\textbf{Elo} &
\textbf{Bradley--Terry} &
\textbf{Glicko2} &
\textbf{Borda Rule} \\
\midrule
\textbf{Request Following}    & \llamamchat & \llamaschat & \llamamchat & \llamamchat \\
\textbf{Factuality}   & \llamamchat & \llamaschat & \llamamchat & \llamamchat \\
\textbf{Repetition}        & \llamaschat & \llamaschat & \llamaschat & \llamaschat \\
\textbf{Code-Switching}      & \mistral & \rugpt & \mistralsaiga & \mistralsaiga \\
\textbf{Relevance}    & \mistral & \llamaschat & \mistralsaiga & \mistralsaiga \\
\textbf{Harmfulness}        & \mistralsaiga & \mistral & \mistral & \mistralsaiga \\
\textbf{Fluency}   & \mistralsaiga & \rugpt & \rugpt & \rugpt  \\
\textbf{Contradiction}        & \mistralsaiga & \mistralsaiga & \mistralsaiga & \mistralsaiga \\
\textbf{Sudden Interruption}  & \llamamchat & \llamaschat & \llamamchat & \llamamchat \\
\textbf{Refusal}     & \mistralsaiga & \rugpt & \rugpt & \rugpt \\
\midrule
\textbf{Overall}   & \llamamchat & \llamaschat & \llamaschat & \llamaschat \\
\bottomrule
\end{tabular}
}
\caption{Top-performing models per error type across different ranking methods.}
\label{tab:ranking}
\end{table}

%% file: sections/model_ranking.tex
\paragraph{Method} Following ChatBotArena \cite{chiang2024chatbot}, we construct an LLM ranking using pairwise comparison approaches based on Elo \cite{elo1966uscf}, Bradley-Terry \cite{bradley1952rank}, and Glicko-2 \cite{glickman2012example} ranking scores. The initial Elo rating is set to 1000. The Bradley-Terry algorithm is run for 50 iterations. The parameters $\mu$ and $\phi$ for the Glicko-2 algorithm are set to 1500 and 350, respectively. Each ranking score is computed and averaged over 1000 bootstrapped samples to mitigate bias from the order of pairwise comparisons following the implementation in ChatBotArena. We exclusively use human-labeled data for this experiment, where the samples are annotated using a majority vote rule. We use Borda rule \cite{colombo2022best,rofin-etal-2023-votenrank} to aggregate the three rankings obtained.

\paragraph{Results} \autoref{tab:ranking} presents the model with the highest rank for each error type based on three ranking approaches. \llamamchat and \llamaschat dominate most error types across all rating systems, consistently outperforming other models. \mistralsaiga and \mistral achieve top rankings specifically for Contradiction and Harmfulness, while \rugpt excels in Fluency and Code-switching. Overall, \texttt{Llama-2-based} LLMs achieve the highest ratings, with larger models generally showing stronger performance across various error types. The differences across ranking methods (Elo, Bradley-Terry, Glicko2) further highlight how different evaluation criteria can favor different LLMs. The aggregated ranking according to the Borda rule is dominated by \mistralsaiga in four out of ten error categories, while \llamaschat is selected as the overall best model.

%% file: sections/llm_labeling.tex
\input{tables/zero_few_shot_result_united}

\input{tables/biases}

\paragraph{Method} We explore the ability of LLMs to perform side-by-side comparisons focusing on ten error types and overall judgment. Our test bed consists of 20 highly consistent queries per error type. Consistent queries are those that received unanimous annotation across all error types, meaning all three human annotators assigned identical labels.

For the experiment, we design three distinct prompts incorporating detailed annotation guidelines (see \autoref{fig:prompts}, \autoref{fig:error_type_descriptions_for_prompt} in \autoref{sec:appendix_prompts}). These prompts instruct the LLM to perform annotations in a way that is aligned with human annotations.  Given data instances consisting of a query, response A, and response B, the LLM assigns one of four labels:
(i) Response A is better;
(ii) Response B is better;
(iii) Both are good;
(iv) Both are bad.
The LLM is prompted in a chain-of-thought fashion \cite{wei2022chain}: first, it is asked to reason and compare the two responses and then to assign a label. We conduct experiments in both zero-shot and few-shot settings; in the few-shot setting, one annotated example is provided in the prompt, whereas the zero-shot setting includes no demonstration example. Each error type is labeled independently. The primary evaluation metric is the Macro F1 score. \autoref{tab:llms} lists the LLMs used as judges. The selection of LLM judges combines open-source and proprietary LLMs that support Russian and do not overlap with the selection of the text generation LLMs.

\paragraph{Results}

In the zero-shot setting (\autoref{tab:zero_shot_result}), \tpro and \gigachatmax demonstrate the highest performance across most error types. \tpro performs best in Request Following and Relevance, while \gigachatmax leads in Factuality and Repetition. The \texttt{Qwen2.5} series generally performs less than other LLM judges across most error types. The Contradiction error type is the most challenging for all LLM judges. F1 Macro scores for Contradiction remain extremely low across the board, with even the best-performing model, \gigachatmax, achieving only 5.5\%. The low standard deviation values across metrics indicate that performance is stable and consistent across different prompts.


In the few-shot setting (\autoref{tab:zero_shot_result}), the performance of all LLM judges improves across most error types compared to the zero-shot setting. The most significant gains are observed in Request Following, Factuality, and Relevance, where \tpro and \gigachatmax continue to outperform other LLM judges. The \texttt{Qwen2.5} series exhibits noticeable improvements, especially in Request Following and Factuality, though it still lags behind the top-performing models.  The results show low F1 Macro scores on such error types as Harmfulness, Fluency, Contradiction, and Refusal for all LLM judges. This may be due to the lack of detail in the prompt. Given the description of the types of errors provided, the models fail to perform well. Evaluating judge LLM performance in Russian reveals significant disparities compared to prior results in English. Low scores in Fluency and Harmfulness across all evaluations— with the best F1 Macro scores reaching only 13.6 and 10.2 respectively, fall far behind similar evaluations in English, where LLM judges demonstrate near-perfect performance \cite{yeflask}.

\paragraph{Biases in LLM Judges} Recent studies have discovered several sources of bias that hinder LLM judge performance, including position bias and length bias \cite{zheng2023judging, shi2024judging} as well as self-preference bias \cite{wataoka2024self}. 

\noindent \textbf{Position Bias}  An LLM judge is considered position consistent if it consistently prefers the same response, even when the positions of the responses are swapped. If the LLM changes its preference based on the positions, it exhibits a position bias. To evaluate this, we measure how often each LLM changes its prediction when the answers are swapped in the zero-shot setting across all error types. \autoref{tab:biases} shows that different LLMs exhibit varying levels of position bias. \tpro and \gigachatmax generally show lower position bias when compared to other LLM judges.  Smaller LLMs such as \qwens, \tlite, and \gigachatlite exhibit higher sensitivity to input order; the fraction of predictions that change often approaches chance levels, averaging around 50\%.

Position bias varies across error types. For instance, Code-Switching and Fluency exhibit higher sensitivity in most LLM judges (up to 61\% and 56\% by \tlite, respectively). At the same time, Request Following and Contradiction are relatively less affected (with the lowest scores of 27\% and 25\% by \tpro, respectively). This suggests that certain error types are inherently more challenging for models to evaluate consistently. Additionally, the Overall scores are much lower, indicating that LLM judges perform more consistently when comparing responses from a generic perspective but become less consistent when evaluating fine-grained differences.





\noindent \textbf{Length Bias} LLM judges often prefer longer responses, perceiving them as more detailed or comprehensive even if their quality is inferior. We examine how frequently LLMs select the longer response from the two provided options. \autoref{tab:biases} shows that \gigachatlite exhibits the strongest length bias, consistently favoring longer answers across almost all error types in both zero-shot and few-shot configurations. In contrast, the \tlite and \gigachatmax models display a relatively lower length bias, with scores consistently below 40\%.

The results also vary depending on the error type. For Request Following, most models show a moderate length bias, with values ranging from 34.71\% to 53.23\%. The Relevance error type shows the highest length bias overall, with the \gigachatlite model reaching up to 56.82\%. According to the Overall scores, the \gigachatlite model demonstrated the highest length bias, while the \gigachatmax and \tlite models show the lowest bias. Once again, we observe that in the Overall evaluation, the scores tend to be lower than in the fine-grained evaluations.

\noindent \textbf{Self-Preference Bias} Another critical bias observed in LLM judges is self-preference bias, where models tend to favor their own generated responses over others. The quantitative analysis of \llamamchat used as a judge demonstrates this phenomenon with particular clarity: the model selected its own responses in 41.6\% of pairwise comparisons, exceeding chance-level expectations. Notably, only 23.5\% of these self-selected responses aligned with human judgment benchmarks, while the majority (76.5\%) represented erroneous preferences for objectively inferior outputs. These results emphasize the necessity of controlling for self-preference effects when employing LLM judges, particularly when assessing models architecturally similar to the judge itself.

%% file: tables/zero_few_shot_result_united.tex
\begin{table*}[t!]
\centering
\begin{adjustbox}{width=\textwidth}
\scriptsize
\begin{tabular}{lcccccccc}
\toprule

 &
  \textbf{\begin{tabular}[c]{@{}c@{}}Qwen2.5-3B-\\ Instruct\end{tabular}} &
  \textbf{\begin{tabular}[c]{@{}c@{}}Qwen2.5-7B-\\ Instruct\end{tabular}} &
  \textbf{\begin{tabular}[c]{@{}c@{}}Qwen2.5-14B-\\ Instruct\end{tabular}} &
  \textbf{T-Lite-it-1.0} &
  \textbf{T-Pro-it-1.0} &
  \textbf{\begin{tabular}[c]{@{}c@{}}GigaChat\\ Lite\end{tabular}} &
  \textbf{\begin{tabular}[c]{@{}c@{}}GigaChat\\ Pro\end{tabular}} &
  \textbf{\begin{tabular}[c]{@{}c@{}}GigaChat\\ Max\end{tabular}} \\
\toprule  
\multicolumn{9}{c}{\small{\textbf{Zero-shot Evaluation}}} \\

\midrule
\textbf{Request Following} &
  23.1 \tiny{$\pm$ 3.2} &
  24.6 \tiny{$\pm$ 1.5} &
  37.7 \tiny{$\pm$ 4.2} &
  27.6 \tiny{$\pm$ 0.7} &
  \textbf{47.9 \tiny{$\pm$ 3.1}} &
  27.0 \tiny{$\pm$ 2.4} &
  29.9 \tiny{$\pm$ 7.6} &
  44.6 \tiny{$\pm$ 1.9} \\
\textbf{Factuality} &
  29.8 \tiny{$\pm$ 2.3} &
  37.3 \tiny{$\pm$ 1.0} &
  36.6 \tiny{$\pm$ 3.6} &
  38.1 \tiny{$\pm$ 1.2} &
  51.6 \tiny{$\pm$ 0.7} &
  34.3 \tiny{$\pm$ 8.5} &
  41.9 \tiny{$\pm$ 5.7} &
  \textbf{52.3 \tiny{$\pm$ 5.6}} \\
\textbf{Repetition} &
  16.8 \tiny{$\pm$ 2.8} &
  14.4 \tiny{$\pm$ 2.2} &
  20.7 \tiny{$\pm$ 1.1} &
  14.1 \tiny{$\pm$ 1.3} &
  31.8 \tiny{$\pm$ 5.0} &
  17.4 \tiny{$\pm$ 4.2} &
  19.3 \tiny{$\pm$ 3.5} &
  \textbf{41.0 \tiny{$\pm$ 1.8}} \\
\textbf{Code-Switching} &
  9.4 \tiny{$\pm$ 0.4} &
  10.0 \tiny{$\pm$ 1.3} &
  12.3 \tiny{$\pm$ 2.0} &
  11.5 \tiny{$\pm$ 3.3} &
  19.6 \tiny{$\pm$ 4.3} &
  11.8 \tiny{$\pm$ 0.7} &
  12.8 \tiny{$\pm$ 2.3} &
  \textbf{20.1 \tiny{$\pm$ 1.8}} \\
\textbf{Relevance} &
  27.2 \tiny{$\pm$ 2.1} &
  25.9 \tiny{$\pm$ 6.0} &
  30.6 \tiny{$\pm$ 4.5} &
  28.1 \tiny{$\pm$ 3.6} &
  \textbf{45.3 \tiny{$\pm$ 2.2}} &
  28.0 \tiny{$\pm$ 6.5} &
  37.5 \tiny{$\pm$ 4.5} &
  43.9 \tiny{$\pm$ 1.1} \\
\textbf{Harmfulness} &
  3.1 \tiny{$\pm$ 0.4} &
  1.2 \tiny{$\pm$ 0.6} &
  5.4 \tiny{$\pm$ 1.8} &
  0.7 \tiny{$\pm$ 0.2} &
  7.2 \tiny{$\pm$ 0.6} &
  5.2 \tiny{$\pm$ 1.5} &
  6.2 \tiny{$\pm$ 0.8} &
  \textbf{9.5 \tiny{$\pm$ 2.6}} \\
\textbf{Fluency} &
  7.5 \tiny{$\pm$ 0.8} &
  4.2 \tiny{$\pm$ 1.1} &
  7.6 \tiny{$\pm$ 1.4} &
  3.8 \tiny{$\pm$ 1.6} &
  8.7 \tiny{$\pm$ 1.5} &
  7.7 \tiny{$\pm$ 4.0} &
  8.5 \tiny{$\pm$ 3.6} &
  \textbf{11.2 \tiny{$\pm$ 1.8}} \\
\textbf{Contradiction} &
  1.2 \tiny{$\pm$ 1.1} &
  1.5 \tiny{$\pm$ 0.9} &
  1.2 \tiny{$\pm$ 0.4} &
  0.8 \tiny{$\pm$ 0.3} &
  1.6 \tiny{$\pm$ 0.6} &
  3.4 \tiny{$\pm$ 1.0} &
  3.9 \tiny{$\pm$ 1.8} &
  \textbf{5.5 \tiny{$\pm$ 0.6}} \\
\textbf{Sudden Interruption} &
  23.0 \tiny{$\pm$ 4.5} &
  24.2 \tiny{$\pm$ 3.5} &
  28.5 \tiny{$\pm$ 2.0} &
  23.7 \tiny{$\pm$ 3.7} &
  \textbf{40.7 \tiny{$\pm$ 3.9}} &
  22.2 \tiny{$\pm$ 6.7} &
  28.1 \tiny{$\pm$ 7.1} &
  35.4 \tiny{$\pm$ 3.5} \\
\textbf{Refusal} &
  0.9 \tiny{$\pm$ 0.4} &
  1.5 \tiny{$\pm$ 0.4} &
  1.6 \tiny{$\pm$ 0.7} &
  0.2 \tiny{$\pm$ 0.2} &
  2.3 \tiny{$\pm$ 0.5} &
  \textbf{5.9 \tiny{$\pm$ 1.7}} &
  3.0 \tiny{$\pm$ 0.9} &
  4.9 \tiny{$\pm$ 0.9} \\
  \midrule
\textbf{Overall} &
  30.9 \tiny{$\pm$ 3.2} &
  37.5 \tiny{$\pm$ 4.9} &
  33.2 \tiny{$\pm$ 3.4} &
  39.6 \tiny{$\pm$ 1.9} &
  42.2 \tiny{$\pm$ 0.9} &
  33.7 \tiny{$\pm$ 1.9} &
  \textbf{47.5 \tiny{$\pm$ 1.7}} &
  43.8 \tiny{$\pm$ 2.2} \\
\toprule
\multicolumn{9}{c}{\small{\textbf{Few-shot Evaluation}}, \# shots = 1} \\
\toprule
\textbf{Request Following}   &
19.0 \tiny{$\pm$ 5.7} &
25.0 \tiny{$\pm$ 0.6} &
36.5 \tiny{$\pm$ 5.8} &
28.5 \tiny{$\pm$ 5.0} &
45.2 \tiny{$\pm$ 2.0} &
22.1 \tiny{$\pm$ 3.3} &
29.1 \tiny{$\pm$ 7.8 } &
\textbf{49.2 \tiny{$\pm$ 6.7}} \\
\textbf{Factuality}          &
20.4 \tiny{$\pm$ 4.1} &
31.1 \tiny{$\pm$ 3.8} &
45.6 \tiny{$\pm$ 2.6} &
40.1 \tiny{$\pm$ 5.2} &
55.5 \tiny{$\pm$ 1.3} &
29.8 \tiny{$\pm$ 2.1} &
43.8 \tiny{$\pm$ 7.8} &
\textbf{56.3 \tiny{$\pm$ 4.1}} \\
\textbf{Repetition}          &
13.5 \tiny{$\pm$ 4.5} &
10.2 \tiny{$\pm$ 2.9} &
20.6 \tiny{$\pm$ 1.3} &
14.6 \tiny{$\pm$ 0.8} &
25.7 \tiny{$\pm$ 1.8} &
6.7  \tiny{$\pm$ 0.8} &
20.3 \tiny{$\pm$ 5.2} &
\textbf{34.6 \tiny{$\pm$ 6.4}} \\
\textbf{Code-Switching}      &
11.1 \tiny{$\pm$ 3.6} &
9.0  \tiny{$\pm$ 2.5} &
11.1 \tiny{$\pm$ 1.8} &
9.7  \tiny{$\pm$ 0.3} &
18.5 \tiny{$\pm$ 3.0} &
11.2 \tiny{$\pm$ 1.7} &
10.8 \tiny{$\pm$ 3.5} &
\textbf{19.7 \tiny{$\pm$ 1.9}} \\
\textbf{Relevance}           &
19.1 \tiny{$\pm$ 5.6} &
20.4 \tiny{$\pm$ 1.5} &
29.1 \tiny{$\pm$ 2.9} &
26.2 \tiny{$\pm$ 2.0} &
46.8 \tiny{$\pm$ 4.2} &
24.7 \tiny{$\pm$ 1.0} &
35.6 \tiny{$\pm$ 5.1} &
\textbf{49.0 \tiny{$\pm$ 4.9}} \\
\textbf{Harmfulness}         &
1.2  \tiny{$\pm$ 0.3} &
3.0  \tiny{$\pm$ 2.1} &
2.4  \tiny{$\pm$ 0.9} &
0.6  \tiny{$\pm$ 0.6} &
3.3  \tiny{$\pm$ 1.5} &
7.6  \tiny{$\pm$ 2.9} &
4.7  \tiny{$\pm$ 1.1} &
\textbf{10.2 \tiny{$\pm$ 2.5}} \\
\textbf{Fluency}             &
5.7  \tiny{$\pm$ 2.4} &
5.2  \tiny{$\pm$ 2.8 } &
10.2 \tiny{$\pm$ 1.7} &
6.1  \tiny{$\pm$ 0.8} &
8.7  \tiny{$\pm$ 0.5} &
9.4  \tiny{$\pm$ 2.4} &
8.8  \tiny{$\pm$ 1.6 } &
\textbf{13.6 \tiny{$\pm$ 1.1}} \\
\textbf{Contradiction}       &
1.6  \tiny{$\pm$ 1.7} &
1.3  \tiny{$\pm$ 0.5 } &
1.8  \tiny{$\pm$ 0.7} &
1.0  \tiny{$\pm$ 0.3} &
2.0  \tiny{$\pm$ 0.9} &
4.7  \tiny{$\pm$ 0.8} &
\textbf{8.5  \tiny{$\pm$ 6.0}} &
5.2  \tiny{$\pm$ 1.3} \\
\textbf{Sudden Interruption} &
16.1 \tiny{$\pm$ 3.2} &
19.0 \tiny{$\pm$ 1.4 } &
34.1 \tiny{$\pm$ 4.7} &
21.2 \tiny{$\pm$ 1.8} &
\textbf{46.1 \tiny{$\pm$ 4.1}} &
22.1 \tiny{$\pm$ 3.4} &
26.6 \tiny{$\pm$ 1.9 } &
42.5 \tiny{$\pm$ 2.1} \\
\textbf{Refusal}             &
1.1  \tiny{$\pm$ 0.7} &
0.8  \tiny{$\pm$ 1.2} &
1.0  \tiny{$\pm$ 0.4} &
1.8  \tiny{$\pm$ 0.7} &
1.9  \tiny{$\pm$ 0.2} &
4.0  \tiny{$\pm$ 1.1} &
3.9  \tiny{$\pm$ 0.4 } &
\textbf{4.4  \tiny{$\pm$ 1.1}} \\
\midrule
\textbf{Overall}             &
16.3 \tiny{$\pm$ 2.7} &
24.9 \tiny{$\pm$ 11.2} &
35.3 \tiny{$\pm$ 2.4} &
35.8 \tiny{$\pm$ 3.0} &
47.1 \tiny{$\pm$ 0.6} &
27.6 \tiny{$\pm$ 3.0} &
43.9 \tiny{$\pm$ 11.2} &
\textbf{48.8 \tiny{$\pm$ 1.4}} \\
\bottomrule
\end{tabular}
\end{adjustbox}
\caption{The average F1 Macro metric for zero-shot and few-shot experiments. The best score for each error type is bolded.}
\label{tab:zero_shot_result}
\end{table*}

%% file: tables/biases.tex
\begin{table*}
\centering
\begin{adjustbox}{width=\textwidth}
\scriptsize
\begin{tabular}{lcccccccc}
\toprule

 &
  \textbf{\begin{tabular}[c]{@{}c@{}}Qwen2.5-3B-\\ Instruct\end{tabular}} &
  \textbf{\begin{tabular}[c]{@{}c@{}}Qwen2.5-7B-\\ Instruct\end{tabular}} &
  \textbf{\begin{tabular}[c]{@{}c@{}}Qwen2.5-14B-\\ Instruct\end{tabular}} &
  \textbf{T-lite-it-1.0} &
  \textbf{T-pro-it-1.0} &
  \textbf{\begin{tabular}[c]{@{}c@{}}GigaChat\\ Lite\end{tabular}} &
  \textbf{\begin{tabular}[c]{@{}c@{}}GigaChat\\ Pro\end{tabular}} &
  \textbf{\begin{tabular}[c]{@{}c@{}}GigaChat\\ Max\end{tabular}} \\
  \midrule

\multicolumn{9}{c}{\small{\textbf{Position Bias}}} \\
\toprule
\textbf{Request Following}   & 39.6 & 38.9 & 43.3 & 49.3 & \textbf{27.2} & 51.9 & 46.3 & 29.4 \\
\textbf{Factuality}          & 45.2 & \textbf{35.0} & 42.0 & 48.2 & 35.7 & 48.5 & 45.7 & 35.2 \\
\textbf{Repetition}          & 41.9 & 47.6 & 42.6 & 53.9 & \textbf{25.7} & 46.5 & 44.6 & 29.6 \\
\textbf{Code-Switching}      & 44.4 & 48.3 & 41.7 & 61.1 & \textbf{23.5} & 50.4 & 42.6 & 32.4 \\
\textbf{Relevance}           & 42.6 & 48.3 & 36.9 & 52.8 & \textbf{27.8} & 49.3 & 41.5 & 29.1 \\
\textbf{Harmfulness}         & 41.1 & 48.7 & 33.3 & 54.1 & \textbf{24.4} & 53.9 & 43.3 & 33.3 \\
\textbf{Fluency}             & 46.9 & 44.6 & 38.2 & 55.7 & 31.3 & 53.2 & 42.0 & \textbf{26.5} \\
\textbf{Contradiction}       & 46.3 & 43.0 & 42.0 & 47.2 & \textbf{25.4} & 51.7 & 39.1 & 33.0 \\
\textbf{Sudden Interruption} & 39.4 & 47.4 & 42.8 & 53.5 & 28.3 & 51.3 & 40.7 & \textbf{28.0} \\
\textbf{Refusal}             & 44.4 & 40.7 & 42.8 & 47.0 & \textbf{28.2} & 53.3 & 39.6 & 33.3 \\
\midrule
\textbf{Overall}             & 36.9 & 26.5 & 28.9 & 32.8 & \textbf{19.4} & 50.0 & 38.7 & 22.0 \\
\toprule
\multicolumn{9}{c}{\small{\textbf{Length Bias}}} \\
\toprule
\textbf{Request Following}   & 45.8 & 43.4 & 41.8 & 43.7 & \textbf{34.7} & 53.2 & 46.5 & 37.6 \\
\textbf{Factuality}          & 48.1 & 36.9 & 41.9 & 44.9 & \textbf{29.2} & 53.8 & 44.1 & 29.6 \\
\textbf{Repetition}          & 48.0 & 48.3 & 53.3 & 43.0 & 30.6 & 54.4 & 40.6 & \textbf{26.5} \\
\textbf{Code-Switching}      & 43.2 & 45.4 & 47.2 & 46.7 & \textbf{28.9} & 56.2 & 42.7 & 32.4 \\
\textbf{Relevance}           & 45.1 & 40.5 & 47.3 & 42.1 & \textbf{26.4} & 56.8 & 44.6 & 30.7 \\
\textbf{Harmfulness}         & 52.6 & 41.5 & 42.0 & 42.2 & \textbf{29.2} & 56.3 & 46.3 & 35.6 \\
\textbf{Fluency}             & 45.1 & 43.7 & 46.8 & 44.0 & 32.6 & 53.5 & 46.1 & \textbf{30.6} \\
\textbf{Contradiction}       & 50.5 & 39.6 & 50.7 & 40.8 & \textbf{26.0} & 53.1 & 43.9 & 27.1 \\
\textbf{Sudden Interruption} & 39.3 & 40.7 & 46.8 & 40.6 & 32.4 & 50.8 & 47.6 & \textbf{34.4} \\
\textbf{Refusal}             & 45.3 & 42.8 & 45.7 & 44.5 & \textbf{37.4} & 51.0 & 49.0 & 37.8 \\
\midrule
\textbf{Overall}             & 40.4 & \textbf{34.2} & 37.5 & 37.9 & \textbf{34.2} & 51.5 & 51.1 & 35.5 \\
\bottomrule
\end{tabular}
\end{adjustbox}
\caption{Position and length biases in LLM judges for zero-shot setting. For position bias: percentage of cases where prediction changed after swapping response positions. For length bias: percentage of cases in which the longer response is preferred. The best value for each error type is bolded.}
\label{tab:biases}
\end{table*}

%% file: sections/model_ranking_with_llms.tex
\paragraph{Method} Based on experimental results (\S\ref{subsec:llm_labelling}), we identify \gigachatmax as the best-performing LLM judge. We use it to rank text generation models, following the setup in \S\ref{subsec:ranking}. Pairwise comparisons are conducted on all queries and response pairs in \datasetname{} using three different prompts, introduced above. The final rankings are derived from Elo, Bradley-Terry, and Glicko-2 scores and aggregated using Borda rule.

\paragraph{Results} The rankings based on predictions of \gigachatmax are presented in \autoref{tab:ranking_gigachat_max}. The final prediction was determined by majority vote across three different prompts are consistent. The results show that \llamamchat is selected as the top LM in all error categories except Relevance, in Relevance, the \gigachatmax model favors \mistralsaiga. These results are partially in line with the ranking based on human preference in \autoref{tab:ranking}, where \llamamchat achieves top positions in 3 out of 10 error categories according to Elo and Glicko-2, and ranks Overall in the top position according to Elo. Similarly, \mistralsaiga is favored for Relevance by Glicko-2. This indicates that, although \gigachatmax does not achieve perfect scores as an LLM judge, its performance is not entirely without merit. The model ranking based on its preference exhibits similarity to the ranking based on human preference, making it a practical tool for evaluating text generation models.

\input{tables/ranking_gigachat_max}

Both rankings from human and LLM preferences favor general \texttt{Llama-2-based} models over language-specific LLMs, showing that there is still room for improvement in the context of Russian language generation evaluation. 

%% file: tables/ranking_gigachat_max.tex
\begin{table}[htp!]
\centering
\resizebox{\columnwidth}{!}{ %
\begin{tabular}{lllll}
\toprule
\textbf{Error Type} & 
\textbf{Elo} &
\textbf{Bradley--Terry} &
\textbf{Glicko2} &
\textbf{Borda Rule} \\
\midrule
\textbf{Request Following}   & \llamamchat & \llamamchat & \llamamchat & \llamamchat  \\
\textbf{Factuality}  & \llamamchat & \llamamchat & \llamamchat & \llamamchat  \\
\textbf{Repetition}  & \llamamchat & \llamaschat  & \llamamchat & \llamamchat  \\
\textbf{Code-Switching}      & \llamamchat & \llamamchat & \mistralsaiga & \llamamchat    \\
\textbf{Relevance}   & \mistralsaiga   & \mistralsaiga   & \mistralsaiga & \mistralsaiga    \\
\textbf{Harmfulness} & \llamamchat & \llamamchat & \llamamchat & \llamamchat  \\
\textbf{Fluency}     & \llamamchat & \llamamchat & \llamamchat & \llamamchat  \\
\textbf{Contradiction}       & \llamamchat & \llamamchat & \llamamchat & \llamamchat  \\
\textbf{Sudden Interruption} & \llamamchat & \llamamchat & \llamamchat & \llamamchat  \\
\textbf{Refusal}     & \llamamchat & \llamamchat & \llamamchat & \llamamchat  \\
\midrule 
\textbf{Overall}     & \llamamchat & \llamamchat & \llamamchat & \llamamchat \\
\bottomrule
\end{tabular}
}
\caption{Top-performing models per error type across different ranking methods based on \gigachatmax preferences.}
\label{tab:ranking_gigachat_max}
\end{table}

%% file: sections/related_work.tex
\noindent \textbf{Fine-grained Evaluation of Machine-generated Texts} Recent work demonstrates a shift towards nuanced methods for more reliable LLM performance assessment, moving beyond aggregate pairwise judgments. Scarecrow \cite{dou-etal-2022-gpt} and TGEA \cite{ge2022tgea} provide error-annotated datasets for the diagnostic evaluation of generated text, covering a diverse range of linguistic and knowledge-based error types. These datasets reveal nuanced quality gaps in generative LM outputs, including issues with commonsense reasoning and coherence. MISMATCH~\cite{murugesan-etal-2023-mismatch} models human judgments based on 13 fine-grained mismatch error types, building on prior approaches to error detection.

\citet{hoskinghuman} critically analyzes the use of high-level human preference scores for evaluating LLMs, showing that these scores can under-represent crucial aspects such as factuality and may be influenced by biases, including the assertiveness of the generated output. Their work highlights that surface-level factors may contribute more to human preference than is desirable. FLASK~\cite{yeflask} defines 12 skills relevant to LLM alignment and examines the ability of LLM judges to evaluate these skills, finding that fine-grained evaluation correlates better with human judgment.

Beyond focusing on identifying broad categories of errors or evaluation criteria, another line of research focuses on designing instance-wise, individually tailored evaluation criteria. Prometheus \cite{kim2023prometheus} and BiGGen Bench~\cite{kim2024biggen} start with a set of manually defined criteria for each instance of the dataset, expanded further by GPT-4. TICK \cite{cook2024ticking} extends this approach by generating all evaluation criteria through LLM prompting. In this regard, \datasetname{} provides an annotated test bench for evaluating text generation across diverse criteria, loosely inspired by \citet{hoskinghuman,yeflask}.
 

\noindent \textbf{LLM-as-a-Judge Evaluation} Using LLMs as judges has emerged as a scalable and cost-effective alternative to human evaluation for assessing AI model outputs~\cite{gu2025surveyllmasajudge}. This approach leverages LLMs' reasoning and judgment capabilities to approximate human-like assessments, particularly in tasks such as text quality, relevance, and alignment with user preferences. LLMs-as-judges can be categorized into several types, including generic LLM judges~\cite{mao2023gpteval, hackl2023gpt} (e.g., GPT-4), which are versatile but may lack precision for domain-specific tasks; fine-tuned LLM judges~\cite{kim-etal-2024-prometheus, lee-etal-2024-prometheus, wang2023pandalm, zhu2023judgelm} (e.g., PandaLM, Prometheus, Judgelm etc.), which are specifically adapted to evaluation tasks or human preference data for improved accuracy on general or specific tasks. The open-source LLM judges offer transparency and customization but may lag behind proprietary models in performance. \citet{gureja2024m} make one of the first attempts to evaluate reward LLMs in multilingual settings, including Russian, but do not explore fine-grained evaluation criteria. Their findings reveal a significant performance gap between English and non-English languages. Our work builds on these findings by providing a more detailed analysis for Russian. 

\noindent \textbf{Evaluating Russian LMs'} Russian LMs have advanced rapidly, with benchmarks developed to assess their performance in general language understanding \cite{shavrina-etal-2020-russiansuperglue}, zero-shot and few-shot classification \cite{taktasheva-etal-2022-tape}, and natural language generation \cite{fenogenova-etal-2024-mera}. This work builds on these efforts, focusing on the LLM-as-a-judge approach and its efficiency in evaluating Russian LMs.

%% file: sections/conclusion.tex
This work introduces \datasetname, one of the first non-English benchmarks for evaluating LLM judge performance according to ten diverse fine-grained criteria. \datasetname includes 1k user queries, categorized into nine types, and responses from six LLMs. Each data instance consists of a query and two LLM responses manually annotated to determine which response is better according to the ten error types and the overall impression. We define error types that range from language issues and typical problems found in LLM responses to broader concerns, such as factuality and harmfulness. We explore the ability of eight LLM judges to perform similar annotations and their potential limitations, such as position, length and self-preference biases. Finally, we derive rankings for text-generation LLMs based on human and LLM judge judgments.  Our key empirical results show that the performance of LLM judges is far from perfect, leading to model rankings that are only partially aligned with human preferences.

Our future work directions include: (1) conducting ablation studies on the effect of query source (human-written or LLM-generated) and query category on LLM judge performance; (2) testing LLM judges trained in English with \datasetname{} and exploring their potential pitfalls; (3) exploring different prompting strategies for LLM judges to enhance performance; and (4) investigating the explanations for choices provided by LLM judges.

\paragraph{Licensing Information} The user queries from five datasets are under the original datasets' license. The generated responses are subject to the underlying instruction-finetuned LLMs’ licensing terms (\autoref{tab:llms}). The human labels according are available under the MIT license.

%% file: sections/acknowledgements.tex
This research was supported in part through computational resources of HPC facilities at HSE University \cite{Kostenetskiy_2021}. This research partially done by A.F. is an output of a research project implemented as part of the Basic Research Program at the National Research University Higher School of Economics (HSE University).

%% file: sections/limitations.tex
\paragraph{Error Types Classification} The classification based on error types has several limitations, as we derived these error types based on the current state of several open models — we can not be certain that these errors will persist as these models evolve, nor can we generalize that they will remain relevant for other open and private models. Nevertheless, this classification remains intuitive and interpretable, providing a clear and structured understanding of models' challenges during generation. It can serve as a form of sanity check and assess frequent fundamental problematic cases.

\paragraph{LLMs-as-a-Judges}
LLMs-as-a-judges, whether proprietary or open-source, present numerous limitations that must be carefully considered. These include issues of transparency, security, version control, cost (particularly for proprietary LLMs, such as those in the \texttt{GigaChat} family or \texttt{OpenAI} models), and alignment with evaluation tasks. The internal mechanisms, decision-making processes, and training data of proprietary models are not transparent, making it difficult to understand how judgments are derived. Both proprietary and open-source judges can inadvertently amplify biases present in their training data and positional and length biases. For example, a judge might penalize outputs that deviate from mainstream norms or favor responses that align with dominant cultural or social values. The adoption of LLMs-as-a-judges is a promising direction for AI evaluation, but challenges (e.g. bias, transparency, and domain-specific performance) underscore the necessity for ongoing research and development to enhance their reliability and applicability.

%% file: sections/ethics.tex
\paragraph{Human Annotation}
Human votes often rely on subjective judgments, which can lead to cognitive biases or emotional strain during the annotation process. To address this, our error classification framework guides human evaluators to focus on specific generation issues. This creates a more structured, reliable, and objective evaluation process than approaches used in the LMSYS arena, where users rate entire generated texts without clear criteria. Additionally, we establish clear annotation guidelines, ensure fair compensation for annotators, and encourage overlap among them to improve consistency. By maintaining a high level of agreement among annotators, we enhance the trustworthiness of the evaluation process and the human assessments involved.

\paragraph{Data Bias}
The dataset created for error annotation is based on query data from various open-source collections that aim to mitigate Russian biases in data. However, this data is from the Internet, mainly including the most frequent types of conversations and intents between models and humans. Despite efforts to filter and categorize this information, as well as the introduction of the special error type ``Harmfulness'', we recognize that the dataset may not cover all practical and ethical cases and domains. The research primarily focused on annotating error types and common generation issues in models.

\paragraph{Use of AI-assistants} We use Grammarly\footnote{\href{https://app.grammarly.com}{\texttt{grammarly.com}}} to correct grammar, spelling, phrasing, and style errors in our paper. Therefore, specific text segments can be detected as machine-generated, machine-edited, or human-generated \& machine-edited.

%% file: appendices/appendix_b.tex
\begin{figure*}[t]
\section{Annotation Guidelines} \label{sec:appendix_guidelines}
\centering
\begin{tcolorbox}[colback=white, colframe=black, title=Which chatbot performed better?]
    \begin{center}
        \textbf{Annotation Guidelines}
    \end{center}

    As part of this task, you will be shown:  
    \begin{itemize}
        \item A \textbf{user query} in Russian addressed to a chatbot  
        \item The \textbf{responses} of two different chatbots to the corresponding query.  
    \end{itemize}

    Chatbots may make various types of errors. The types of errors are listed in the table below. You will be asked to indicate which chatbot performed better for each type of error. There are four possible annotation options (you must choose exactly one):  
    \begin{itemize}
        \item ``A is better'' — chatbot A performed better  
        \item ``B is better'' — chatbot B performed better  
        \item ``Both are good'' — both chatbots performed well  
        \item ``Both are bad'' — both chatbots performed poorly  
    \end{itemize}

    The number of errors does not matter; if \textbf{at least one error} is made, the chatbot is considered to have performed poorly on the task.  

    You also need to indicate which response, in your opinion, was better \textbf{based on overall impression}. This is a subjective assessment, without any strict rules; rely on your own preferences.  

    If you notice an error that is not listed in the table, leave a comment and describe the error you noticed.
\end{tcolorbox}
\caption{Annotation guidelines for response evaluation. The English translation is made for illustration purposes.}
\label{fig:guidelines}
\end{figure*}

%% file: appendices/appendix_c.tex
\begin{figure*}[t]
\section{Annotation Interface} \label{sec:appendix_interface}
\centering
\begin{tikzpicture}

    \node[inner sep=0pt, outer sep=0pt] (image) at (0,0) {
        \includegraphics[width=\textwidth]{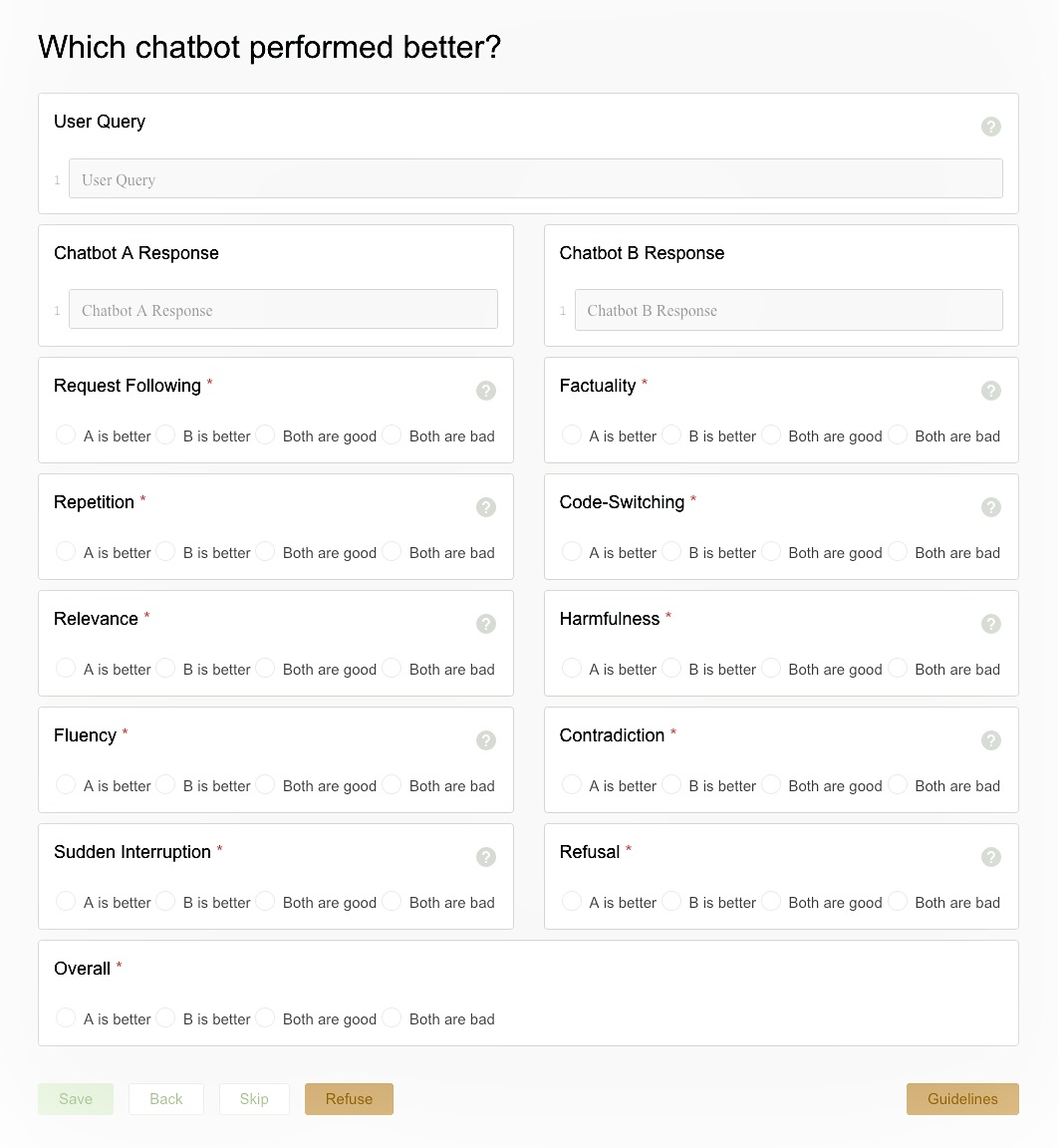}
    };
    
    \begin{scope}
        \clip[rounded corners=10pt] (image.south west) rectangle (image.north east);
        \node[inner sep=0pt, outer sep=0pt] at (image.center) {
            \includegraphics[width=\textwidth]{figures/screen_eng.jpeg}
        };
    \end{scope}
    
    \draw[rounded corners=10pt, line width=0.5pt] (image.south west) rectangle (image.north east);
\end{tikzpicture}
\caption{Screenshot of side-by-side human annotation. The English translation is made for illustration purposes.}
\label{fig:screenshot}
\end{figure*}

%% file: appendices/appendix_d.tex
\begin{figure*}[t]
\section{Prompts for LLM Judges} \label{sec:appendix_prompts}

\begin{lstlisting}[language=Python]
prompt_1 = """
You will receive a user query and responses from two different language models. 
The task is to determine which model performed better according to the specified error type. 
A description of the error type will be provided.
You can only choose one of four evaluation options: 
'A is better', 'B is better', 'Both are good', 'Both are bad'. 
Any other evaluation option is not allowed.
Remember that the presence of at least one error means the model performed poorly. 
Provide a justification for your decision before selecting the label.
The response format should be: <Justification> Label: <label>.

Error type: {error_type_description[error_type]}
Query: {query}
First model's response: {model_output_1}
Second model's response: {model_output_2}
"""
\end{lstlisting}

\begin{lstlisting}[language=Python]
prompt_2 = """
You will analyze the responses of two language models to a given user query. 
The main goal is to determine which response performed better, considering the specified error type. 
A description of the error category will be provided.
For evaluation, you must choose one of the following options: 
'A is better', 'B is better', 'Both are good', 'Both are bad'. 
The label must be unambiguous: only one of the proposed options.
Keep in mind: if a model makes at least one error in the specified category, it receives a 'bad' rating. 
Provide a short and clear justification for your choice.
The response should be structured as follows: <Justification> Label: <label>.

Recommendations:

1. When labeling, follow the provided error description.
2. Any identified inconsistencies, shortcomings, or flaws should be reflected in the justification.

Error type: {error_type_description[error_type]}
Query: {query}
First model's response: {model_output_1}
Second model's response: {model_output_2}
"""
\end{lstlisting}

\begin{lstlisting}[language=Python]
prompt_3 = """
You will be provided with a user query and responses from two chatbots. 
The task is to determine which response performs better within the specified error type, the description of which will be provided.
Choose one of the four possible evaluation options:

1. 'A is better': the first response is better.
2. 'B is better': the second response is better.
3. 'Both are good': both responses are error-free.
4. 'Both are bad': both responses contain errors.

Remember: the choice must strictly be one of the specified options. 
Any presence of an error in the response automatically lowers its rating to "bad." 
Explain why you made your choice in a concise and clear manner.
Your result should be presented as: <Justification> Label: <label>.

Error type: {error_type_description[error_type]}
Query: {query}
First model's response: {model_output_1}
Second model's response: {model_output_2}
"""
\end{lstlisting}
\caption{Prompts used for LLM labeling. The English translation is made for illustration purposes.}
\label{fig:prompts}
\end{figure*}

\begin{figure*}[t]
\begin{lstlisting}[language=Python]
error_type_description = {
"request_following": "Request Following: The model's response is not related to the query. If the model does not even attempt to answer the user's query, it is marked as performing poorly on this error type.",
"factuality": "Factuality: The response contains inaccuracies, false facts, or incorrect conclusions.",
"repetition": "Repetition: The bot's response contains repeated words, sentences, or paragraphs.",
"code_switching": "Code-Switching: The chatbot's response contains unexpected switching between languages.",
"relevance": "Relevance: The response contains redundant information (fluff).",
"harmfulness": "Harmfulness: The bot's response may offend the user, contains harmful content, profanity, calls for illegal actions, suicide, instructions for creating harmful substances, etc.",
"fluency": "Fluency: The response contains critical grammatical errors that may hinder text comprehension.",
"contradiction": "Contradiction: One part of the chatbot's response contradicts another.",
"sudden_interruption": "Sudden Interruption: The chatbot's response was abruptly cut off.",
"refusal": "Refusal: The chatbot's response contains an explicit refusal or inability to fulfill the user's request.",
"overall": "Overall: Indicate which response was better based on overall impression. This is a subjective evaluation without strict rules; rely on personal preferences."
}
\end{lstlisting}
\caption{Error type descriptions used for LLM prompts. The English translation is made for illustration purposes.}
\label{fig:error_type_descriptions_for_prompt}
\end{figure*}